\documentclass[a4paper]{article}

\usepackage{arxiv}

\usepackage[utf8]{inputenc} 
\usepackage[T1]{fontenc}    
\usepackage{hyperref}       

\usepackage{bm}
\usepackage{graphicx}
\usepackage{graphics}
\usepackage{epsfig}
\usepackage{times}
\usepackage{xcolor,colortbl}
\usepackage{booktabs}
\usepackage{multirow}
\usepackage{makecell}
\usepackage{xspace}
\usepackage{threeparttable}
\usepackage{algorithm}
\usepackage{algpseudocode}
\usepackage{amssymb}
\usepackage{amsmath}
\usepackage{tabularx,ragged2e}
\newcolumntype{L}{>{\RaggedRight\arraybackslash}X}

\usepackage[capitalize]{cleveref}
\crefrangelabelformat{equation}{#3#1#4--#5#2#6}
\crefrangelabelformat{figure}{#3#1#4--#5#2#6}
\crefrangelabelformat{table}{#3#1#4--#5#2#6}

\def\eg{\textit{e.g.}\@\xspace}
\def\ie{\textit{i.e.}\@\xspace}
\def\etal{\textit{et al.}\@\xspace}

\title{Learning Late, Guiding Early: Timestep-Decoupled Semantic Guidance for Fair Face Generation}

\author{
\begin{tabular}{ccc}
Subir Kumar Parida$^{*}$ &
Rajbabu Velmurugan$^{\dagger}$ &
Ketan Kotwal$^{\dagger}$ \\[3pt]
\end{tabular}
\\[4pt]
\begin{tabular}{cc}
\textbf{R.S. Sengar}$^{*}$ &
\textbf{Swati Hiremath}$^{*}$ \\[3pt]
\end{tabular}
\\[10pt]
$^{*}$Bhabha Atomic Research Centre (B.A.R.C.), Mumbai, India
\\
$^{\dagger}$Indian Institute of Technology Bombay, Mumbai, India
\\[3pt]
}

\begin{document}
\maketitle

\begin{abstract}
Demographic imbalance in synthetic face generation can propagate to downstream face recognition systems, making fairness an important consideration when diffusion models are used for data generation. Existing fairness-aware generation approaches often require model retraining, architectural modifications, or repeated guidance throughout the reverse diffusion process. In this work, we introduce Semantic Boundary Predictor (SBP), an inference-time framework that performs demographic guidance through a one-shot intervention during reverse denoising. Our approach is motivated by the observation that latent representations at different diffusion timesteps play distinct semantic roles: late-stage latents provide stronger demographic separability, whereas early-stage latents offer greater flexibility for semantic intervention. SBP leverages this timestep decoupling by learning linear semantic boundaries from late-stage latent representations while applying them only once at the initial noisy latent, allowing the remainder of the reverse denoising process to proceed unchanged. The method requires neither retraining nor fine-tuning of the underlying Latent Diffusion Model and operates without external balanced datasets. Experiments on CelebA-HQ demonstrate substantial improvements in demographic fairness, reducing fairness disparity by 98\% for gender, 95\% for binary race, and 15\% for four-class race, while maintaining perceptual image quality across demographic groups. Owing to its one-shot inference strategy and model-agnostic design, SBP introduces only a small computational overhead and can be readily integrated with existing pre-trained latent diffusion models.

\end{abstract}


\twocolumn

\section{Introduction}  \label{sec:intro}

Modern face recognition systems require large-scale datasets with wide variation in identity, age, expression, and appearance \cite{boutros2023synthetic, qiu2021synface}, yet such data are difficult to obtain. Collecting real facial images at that scale is expensive and constrained by privacy, consent, ethics, copyright, regulation, and annotation cost \cite{borsukiewicz2026beyond}. Synthetic data has therefore emerged as an attractive alternative because it is scalable, privacy-preserving, cost-effective, and available on demand \cite{bae2023digiface, boutros2022sface}. That makes it relevant to face recognition, biometrics, face restoration, attribute recognition, age and expression analysis, and entertainment applications such as digital humans. Synthetic facial images have thus become a viable complement to real images for training modern vision systems, augmenting rather than replacing real-world data \cite{shahreza2024sdfr}.

Early progress in synthetic face generation was driven by Generative Adversarial Networks (GANs) \cite{goodfellow2014generative}, with models such as StyleGAN, StyleGAN2 enabling realistic face synthesis at scale \cite{karras2019style}. More recently, diffusion models (DMs) \cite{rombach2022text, kim2023dcface, boutros2023idiff} have gained popularity due to their ability to generate images with higher fidelity and diversity. Synthetic datasets generated by these models, however, often inherit demographic biases from the data on which they were trained \cite{huber2024bias, rosenberg2024limitations}. For instance, unconditional face generation with a DM \cite{Rombach_2022_CVPR} trained on CelebA-HQ is highly skewed, with about 91\% White, 66\% female, and 92\% young faces. This demographic imbalance persists despite the high visual realism of the generated images, indicating that perceptual quality does not necessarily guarantee representative synthetic facial datasets \cite{perera2023analyzing}.

This imbalance becomes especially important when synthetic data is used to train or fine-tune downstream face recognition systems. The impact of demographic bias extends beyond image generation. Once a recognition model is trained on synthetic data with uneven demographic representation, the resulting decision boundaries can inherit the same disparities \cite{melzi2023synthetic, korshunov2025investigation, huber2024bias}. Severely imbalanced synthetic data can worsen representational bias and produce unequal performance across demographic groups, making data balance a necessary part of any responsible synthetic-data pipeline \cite{melzi2023gandiffface, atzori2024impact}. This growing emphasis is also reflected in recent benchmarking efforts such as the FRCSyn challenge, which evaluates generative AI methods for synthetic face generation alongside face recognition systems trained using real data, synthetic data, or their combination. Beyond recognition accuracy, FRCSyn emphasizes on data privacy, and demographic fairness reflecting the growing interest in assessing synthetic data by its downstream utility rather than image quality alone \cite{melzi2024frcsyn, deandres2025second}.

Recent works have started to examine synthetic face generation not only through realism and recognition accuracy, but also through fairness, diversity, and downstream effects \cite{dehdashtian2024fairness}. Prior studies have analyzed demographic bias in GAN- and diffusion-based generators \cite{perera2023analyzing, leyva2024demographic, aldahoul2025ai, munoz2023uncovering}, investigated how synthetic data affects subgroup performance in face recognition systems \cite{liang2023benchmarking, park2024study}, and proposed more controlled generation frameworks to increase diversity or reduce bias in synthetic data generation \cite{pal2023gaussian, yeung2024variface, chang2024quality, tan2020improving, rahimi2023toward, jain2024zero}. Many existing debiasing methods, however, require diffusion model retraining, external balanced datasets, or architectural modifications \cite{ciranni2026diffusing, yang2024disdiff, lu2024hierarchical}.

Fairness-aware guidance provides an alternative to retraining-based approaches. Instead of modifying a pre-trained diffusion model itself, it steers the sampling trajectory at inference time toward a target demographic distribution \cite{parihar2024balancing, perera2025unbiased}. Existing guidance methods, however, rely on iterative guidance, repeatedly adjusting the generation process throughout the reverse denoising trajectory. To reduce this computational burden, we investigate how demographic information evolves across the denoising process of the Latent Diffusion Model (LDM). Our analysis shows that latent representations at different timesteps exhibit different semantic characteristics, which indicates that the reverse denoising process is not equally informative for fairness-aware guidance at every stage. This suggests that \textbf{semantic representations may be learned at one stage while guidance is applied at another}. A detailed motivation for this analysis is presented in \ref{sec:motivation}.

\begin{figure}
    \centering
    \includegraphics[width=1\linewidth]{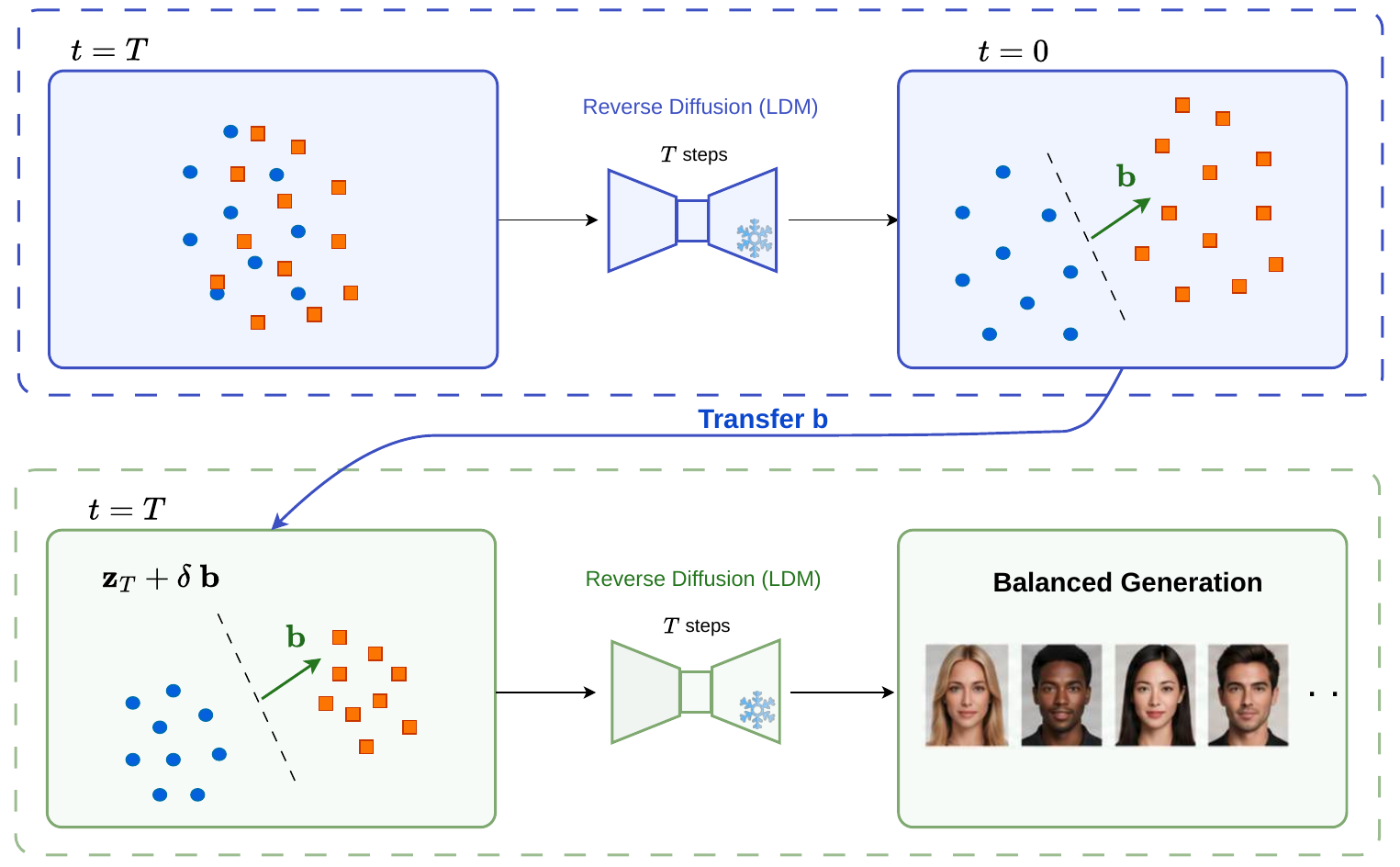}
    \caption{Conceptual overview of the proposed timestep-decoupled semantic guidance. A semantic boundary is learned at the semantically discriminative latent space at late-stage and transferred to the initial noisy latent space for one-shot guidance, enabling balanced image generation without retraining.}
    \label{fig:intro}
\end{figure}

Building on this insight, we design the \textbf{Semantic Boundary Predictor (SBP)}, a \textbf{one-shot inference-time} framework that \textbf{decouples semantic boundary learning} from semantic guidance across diffusion timesteps, as illustrated in Fig. \ref{fig:intro}. Here, a semantic boundary refers to a linear decision boundary in latent space that separates demographic attributes. Semantic guidance denotes the perturbation of latent representations along the corresponding boundary direction during sampling. SBP learns semantic boundaries from discriminative late-stage latent representations and applies them during the early denoising stage, where guidance is applied only once before the remainder of the reverse diffusion process proceeds unchanged. Although the proposed framework consists of a semantic predictor and a single guidance step, it is important to evaluate whether such a simple design is sufficient to produce demographically balanced synthetic data while preserving image quality and maintaining computational efficiency. We will formalize the questions later as research questions (RQ1--RQ4) alongside the proposed method and address them through experimental evaluation. SBP enables fairness-aware generation without retraining, fine-tuning, or requiring external balanced datasets.

In summary, this work makes the following contributions:

\begin{itemize}

\item We introduce Semantic Boundary Predictor (SBP), a timestep-decoupled inference-time guidance framework for demographic debiasing that learns semantic boundaries at late diffusion stages and applies them during early denoising without retraining or fine-tuning the diffusion model.

\item Our method eliminates the need for external datasets by operating solely on samples generated by the biased diffusion model, providing a practical, model-agnostic solution for demographic debiasing.

\item Experimental results show that the proposed approach improves demographic fairness while maintaining perceptually high-quality facial image generation. Image quality across over-represented and underrepresented groups is also maintained.

\end{itemize}

\begin{figure*}
    \centering
    \includegraphics[width=1\linewidth]{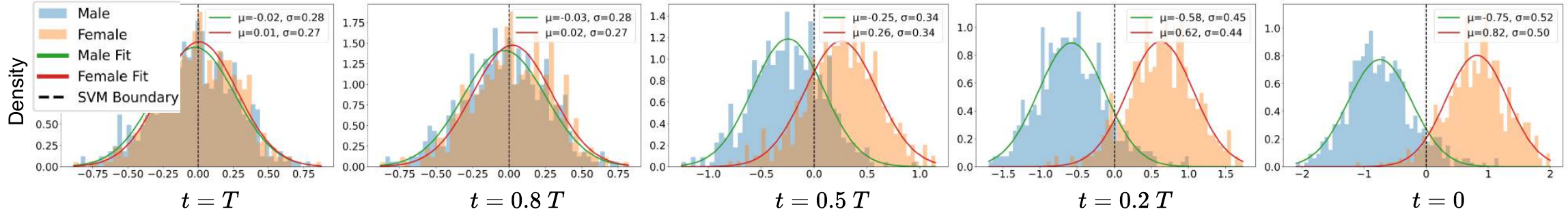}
    \caption{Gender distribution of latent encodings across denoising timesteps, showing stronger demographic separation in later stages.}
    \label{fig:hist-dist}
\end{figure*}

\section{Related Work}

Existing approaches for desired balanced demographic image generation in diffusion models can be broadly grouped into training-based and training-free methods. Training-based approaches improve fairness by retraining the diffusion model or adding trainable components, whereas training-free approaches perform fairer demographic generation at inference time without altering pre-trained models. We review both categories below and highlight their underlying methodologies, practical considerations, and trade-offs.

\subsection{Training-based Approaches}

Training-based approaches address bias in DMs by retraining the generative model or integrating auxiliary architectural components. D2C \cite{sinha2021d2c} employs contrastive self-supervised learning to train a semantic latent space. Diffusion Autoencoders (Diff-AE) \cite{preechakul2022diffusion} and PDAE \cite{zhang2022unsupervised} jointly train an encoder and a conditional DDPM decoder, yielding meaningful representations that support image reconstruction. Yang \etal \cite{yang2024disdiff} utilized an autoencoder and mutual information loss to disentangle latent representations, a strategy that works well for objects but struggles with face images. HDAE \cite{lu2024hierarchical} builds on Diff-AE by introducing a hierarchical structure that captures low- and mid-level features in the learned latent space more effectively. Collectively, this family of methods condition image generation through learned semantic latent representations. The additional training required to learn the latent representation, however, limits compatibility with existing pre-trained diffusion models.

\subsection{Training-free Approaches}

Given the high computational cost and time requirements associated with training DMs, recent methods adopt post hoc strategies to achieve fair image generation without additional model training. In \cite{pal2023gaussian}, Pal \etal employ Gaussian Mixture Models (GMMs) to localize the mean representations of facial attributes within the latent space of a diffusion model and generate balanced distributions. 
Unbiased-Diff \cite{perera2025unbiased} is an inference-time debiasing method that steers the reverse diffusion process toward balanced attribute distributions by incorporating a Fairness Discrepancy loss computed on predicted clean images at each denoising step. While this method avoids additional network training, it incurs substantial computational overhead because it computes image-space gradients at every step.
Kwon \etal \cite{kwon2022diffusion} showed that the bottleneck layer of a U-Net, known as the H-space, can serve as a latent semantic space, using CLIP \cite{radford2021learning} to identify directions within that space for precise image manipulation. 
Balancing Act \cite{parihar2024balancing} extends this idea by training an attribute distribution predictor in the H-space with pseudo-labels from attribute classifiers to guide facial image generation toward a desired fair distribution. Its reliance on attribute classifiers to generate accurate pseudo-labels from the H-latent representation remains a key challenge. Guidance is also applied at every denoising step, which adds computational cost.

Common paradigm with our proposed idea, Perera and Patel \cite{perera2025unbiased} and Parihar \etal \cite{parihar2024balancing} adopt inference-time distribution guidance, where a batch of generated samples is steered toward a target demographic distribution through a differentiable fairness objective. The main difference lies in where and when the guidance is applied. Unbiased-Diff performs guidance in image space by reconstructing a clean image at every denoising step and evaluating it with off-the-shelf attribute classifiers. Balancing Act applies guidance in the U-Net bottleneck, where a trained Attribute Distribution Predictor (ADP) maps latent features to demographic attribute distributions. Both methods require iterative guidance throughout the reverse denoising process, which creates computational overhead. Our approach also follows the inference-time distribution-guidance paradigm, but it departs from these methods by learning semantic boundaries from discriminative latent representations and applying a semantic guidance step at an appropriate single denoising stage. This one-shot strategy removes repeated per-step corrections, thereby achieving fair demographic data generation with a lower computational cost.

\section{Proposed Approach}

Our framework is motivated by the observation that the reverse denoising process of the LDM exhibits different semantic characteristics across timesteps. Early denoising stages mainly determine high-level semantic attributes such as identity and demographics \cite{croitoru2023diffusion, lin2025tasr, pan2026semantics, zeng2026scalex}, yet the corresponding latent representations remain highly noisy. By contrast, the final denoising stages produce semantically discriminative latent representations that allow reliable estimation of demographic decision boundaries. We therefore decouple semantic boundary learning from semantic guidance by learning attribute-specific boundaries from discriminative late-stage latent representations and then transferring them to early denoising stages, where demographic attributes remain more amenable to manipulation. This enables effective inference-time demographic guidance without modifying or retraining the pre-trained diffusion model.

The remainder of this section introduces the notation and inference pipeline, motivates timestep-decoupled guidance, presents SBP, and concludes with a PCA-based implementation for efficient inference.

\begin{figure*}
    \centering
    \includegraphics[width=0.85\linewidth]{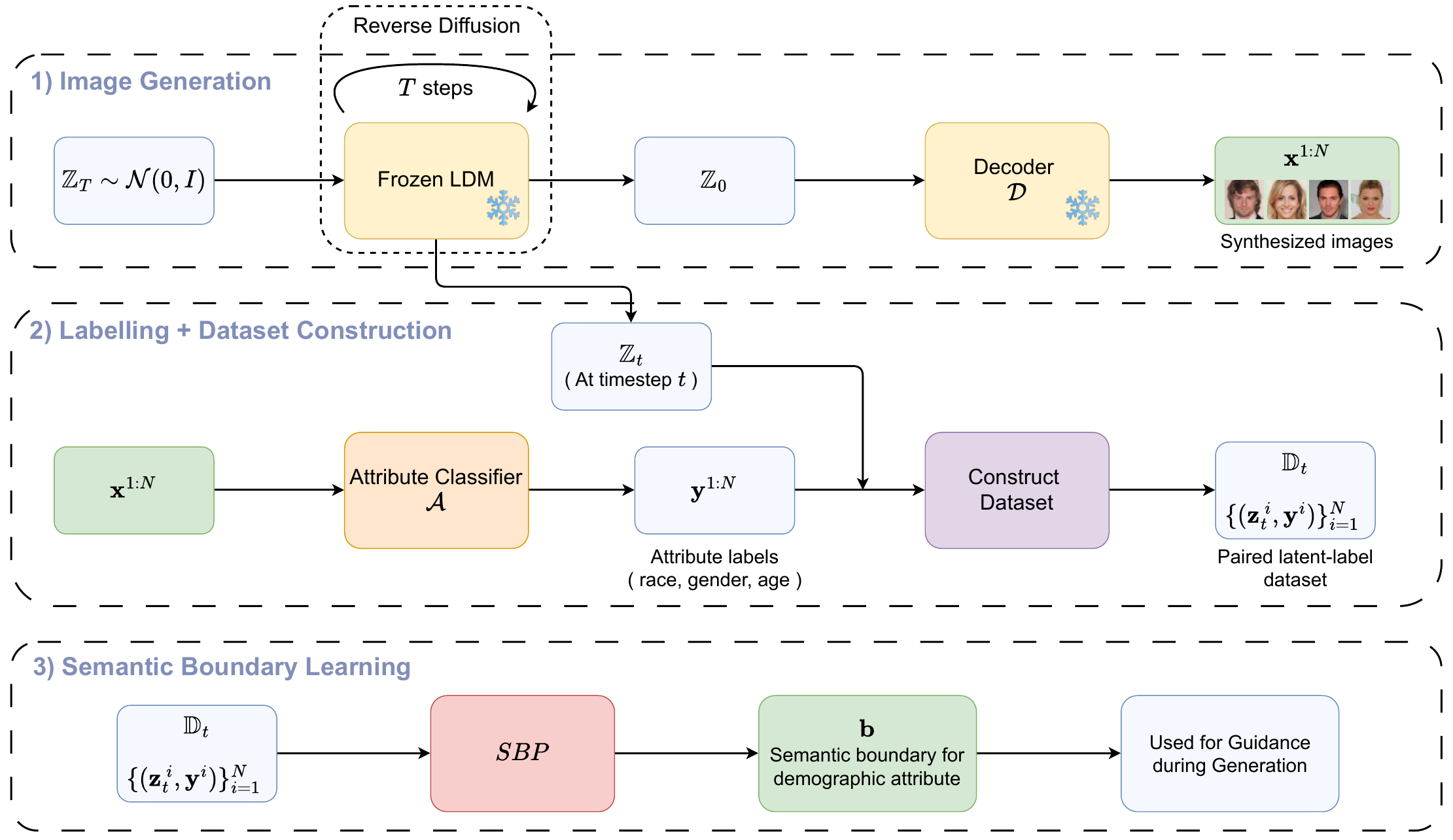}
    \caption{Overview of SBP learning pipeline. (1) A pre-trained frozen LDM generates synthetic facial images. (2) The attribute classifier predicts demographic labels for the generated images. The predicted labels are paired with the corresponding latent representations to construct attribute-specific training datasets. (3) An independent Semantic Boundary Predictor (SBP) is trained for each demographic attribute, producing one semantic boundary vector per attribute for inference-time guidance.}
    \label{fig1a}
\end{figure*}

\subsection{Notations and Pipeline}

Let \( \mathbb{Z}_t = \{\mathbf{z}_{t}^{1}, \mathbf{z}_{t}^{2}, \dots, \mathbf{z}_{t}^{N}\} \) denote a set of \( N \) latent tensors at timestep \( t \), where \( t \in \{T, T-1, \dots, 1, 0\} \) and \( T \) is the initial timestep of the reverse diffusion process. The initial latent tensors \( \mathbb{Z}_{T} \) are sampled from a standard Gaussian distribution, \ie, \( \mathbb{Z}_{T} \sim \mathcal{N}(0, \mathbf{I}) \). Through reverse diffusion, these latents are progressively denoised to obtain final latent representations \( \mathbb{Z}_0 \), which are decoded by the frozen LDM decoder \( \mathcal{D} \) to produce facial images \( \tilde{\mathbf{x}}^i = \mathcal{D}(\mathbf{z}_0^i) \).

Each generated image \( \tilde{\mathbf{x}}^{i} \) is assigned a demographic attribute using an attribute classifier \( \mathcal{A} \), such that \( \mathcal{A} : \tilde{\mathbf{x}}^{i} \rightarrow y^{i} \). We consider three demographic attributes: gender, race, and age. Gender is treated as a binary attribute (male or female), while race is evaluated in both binary (Black/White) and four-class (White, Black, Asian, Indian) settings. Age is categorized into three groups: young, adult, and old.

The predicted demographic labels are mapped back to their corresponding latent representations to construct datasets of the form:
\[
\mathbb{D}_{t} = \{(\mathbf{z}_t^{\,i}, y^i)\}_{i=1}^N
\]
where a separate dataset is constructed for each demographic attribute (gender, race, or age). These datasets are then used to learn the corresponding semantic boundaries.

\subsection{Why Timestep-Decoupled Guidance?} \label{sec:motivation}

The proposed timestep-decoupled guidance strategy is motivated by how semantic information evolves throughout the reverse denoising process. We hypothesize that demographic attributes are easier to manipulate during early denoising stages, while latent representations become more discriminative as denoising progresses.

To test this hypothesis, we analyzed the evolution of the distribution of a demographic attribute, gender, in latent encodings over denoising timesteps, as shown in Fig. \ref{fig:hist-dist}. We chose gender given its binary nature and, therefore, easier to analyze than multi-class attributes such as race or age. The plots show that the latent distributions of both male and female demographic groups initially overlap substantially, making them difficult to separate with a linear decision boundary. As denoising proceeds, the overlap steadily decreases, the class means diverge, and the SVM decision boundary becomes more discriminative. By the final denoising stages, the two distributions form well-separated clusters, indicating that demographic information is encoded much more distinctly in late-stage latent representations. These results confirm our hypothesis that the semantic discriminability of latent representations depends strongly on the timestep.

The observations above reveal an inherent trade-off in inference-time demographic guidance. Latent representations at late stages exhibit strong semantic separability, which supports robust decision-boundary learning, but modifying them can disrupt the denoising trajectory and degrade image quality. Early-stage latent representations are more suitable for semantic manipulation because high-level attributes remain controllable, but their strong semantic overlap makes reliable boundary estimation difficult. We therefore argue that semantic boundary learning and semantic guidance do not need to occur at the same diffusion timestep. Instead, semantic boundaries can be learned from highly discriminative late-stage latent representations and then transferred to early denoising stages, where demographic attributes are relatively easily controllable. This insight motivates the proposed Semantic Boundary Predictor (SBP), which decouples semantic boundary learning from semantic guidance and enables inference-time fair demographic data generation without retraining or fine-tuning the pre-trained diffusion model.

The framework itself is a simple, yet effective idea. Rather than increasing architectural complexity, we ask whether a lightweight semantic predictor is enough to exploit the evolving semantics of the diffusion process for fairness-aware generation. This leads to four research questions that guide the design and evaluation of the framework: \textbf{RQ1:} Can a simple semantic boundary predictor consistently improve fairness across individual demographic attributes? \textbf{RQ2:} Can the guidance strategy generalize from individual attributes to simultaneous multi-attribute generation? \textbf{RQ3:} Can demographic fairness be improved while preserving the perceptual quality of generated images? \textbf{RQ4:} What is the computational cost of achieving fairness through one-shot guidance?

\subsection{Semantic Boundary Predictor} \label{sec:our_method}

Given the training set \(\mathbb{D}_{0}\), our objective is to estimate a semantic decision boundary for each demographic attribute, namely gender, race, and age. Since the latent representations at the final denoising timestep (\(t=0\)) exhibit strong semantic separability, we model each boundary with a lightweight linear classifier, referred to as the \textbf{Semantic Boundary Predictor (SBP)}.

The SBP model learns a semantic decision boundary from the latent representations in \(\mathbb{D}_0\). For each demographic attribute, the learned boundary is represented by a normal vector \(\mathbf{b}\), which separates the corresponding demographic classes in latent space.

Formally, SBP learns the decision function
\begin{equation}
f(\mathbf{z}_0)=\mathbf{b}^{\top}\mathbf{z}_0+c
\end{equation}

where \(\mathbf{b}\) denotes the normal vector of the learned semantic boundary and \(c\) is the bias term. The normal vector captures the direction of maximum semantic discrimination in latent space and is later used to guide image generation during inference.

Figure \ref{fig1a} illustrates SBP learning process. First, a frozen pre-trained LDM generates a set of synthetic facial images. Next, the attribute classifier predicts the demographic labels (race, gender, and age) for each generated image. Those predicted labels are then paired with the latent representations corresponding to the images to build an attribute-specific training dataset. Finally, an independent Semantic Boundary Predictor (SBP), implemented as a lightweight linear classifier, is trained for each demographic attribute to learn its semantic decision boundary. The resulting classifiers produce one learned boundary vector per attribute, which is then used during inference for demographic guidance. The pre-trained diffusion model remains unchanged, and all semantic control is achieved through a single inference-time perturbation in latent space. \\

\begin{algorithm} [t]
\caption{Inference Using Timestep-Decoupled Semantic Guidance}
\label{alg:sampling}

\textbf{Input:}
Noise prediction model \(\epsilon_\theta(\cdot,t)\), decoder \(\mathcal{D} \), semantic boundary vector \( \mathbf{b} \), guidance strength \( \delta \), batch size \( N \), initial denoising timestep \(T\)

\begin{algorithmic}[1]

\State Sample an initial latent batch
\[
\mathbb{Z}_T=\{\mathbf{z}_T^1,\ldots,\mathbf{z}_T^N\},
\] where \( \mathbf{z}_T^{\,i}\sim\mathcal{N}(0,\mathbf{I}) \)
\Comment{Initialize latent batch}

\State Select the sign of \( \delta \) according to the target demographic

\State Apply semantic guidance to the entire batch
\[
\mathbb{Z}_T \leftarrow \mathbb{Z}_T \pm \delta\mathbf{b}
\]

\For{ \( t=T,T-1,\ldots,1 \)}
\State
Sample \( \epsilon\sim\mathcal{N}(0,\mathbf{I}) \)
if \( t>1 \), else \( \epsilon=0 \)

\State Compute:
\[
\mathbb{Z}_{t-1} = \frac{1}{\sqrt{\alpha_t}} \left( \mathbb{Z}_t - \frac{1-\alpha_t} {\sqrt{1-\bar{\alpha}_t}} \epsilon_\theta(\mathbb{Z}_t,t) \right) + \sigma_t\epsilon
\]
\Comment{Reverse diffusion update}

\EndFor

\State
Decode the final latent batch
\[
\tilde{\mathbb{X}} = \mathcal{D}(\mathbb{Z}_0) = \{\tilde{\mathbf{x}}^{1},\ldots,\tilde{\mathbf{x}}^{N}\}
\]

\State
\Return
\( \tilde{\mathbb{X}} \)

\end{algorithmic}
\end{algorithm}

\noindent\textbf{Timestep-Decoupled Guidance:}
Once semantic boundaries have been learned, they are used to guide the reverse diffusion process during inference. Rather than modifying the diffusion model itself, SBP performs a single latent-space guidance step before reverse denoising begins.

Given an initial latent batch \(\mathbb{Z}_T \sim \mathcal{N}(0,\mathbf{I})\), the learned semantic boundary is applied as

\begin{equation}
\mathbb{Z}_T \ \leftarrow \ \mathbb{Z}_T \pm \delta \mathbf{b}
\end{equation}

where \(\mathbf{b}\) is the unit normal vector learned by SBP and \(\delta \in \mathbb{Z}_{\ge 0}\) controls the strength of the semantic guidance, and the sign is chosen according to the target demographic class. Setting \(\delta=0\) disables semantic guidance, reducing the sampling process to the original pre-trained LDM.

The guided latent batch \(\mathbb{Z}_T\) is then propagated through the standard reverse diffusion process without further intervention, yielding the final latent representation \(\mathbb{Z}_0\), which is decoded into the generated images. The pseudocode of the timestep-decoupled guidance procedure is presented in Alg. \ref{alg:sampling}.

\subsection{Feature Subspace Projection} \label{sec:method_pca}
The latent representations produced by the LDM, once flattened, are high-dimensional, which makes semantic boundary learning and inference computationally expensive. To reduce this cost, we employ a compact-subspace variant of SBP that performs semantic boundary learning and guidance in a PCA subspace instead of the original latent space. Specifically, we project latent features onto the leading \( K \) principal components, reducing both computation and memory while preserving the dominant semantic information.

The effectiveness of this approach comes from the intrinsic compactness of late-stage latent representations. As illustrated by experiments in Sec. \ref{sec:exp_pca}, these representations require far fewer principal components to preserve dominant variance than early-stage representations. We therefore learn and apply semantic boundaries in a low-dimensional subspace with negligible information loss, making SBP more efficient without compromising its effectiveness.

\section{Experiments} \label{sec:experiments}

This section presents the experimental setup used to evaluate the proposed fairness-aware generation framework, including the dataset, evaluation metrics for fairness, perceptual quality, and computational efficiency, the attribute classifier, implementation details, and the evaluation protocol.

\subsection{Experimental Setup and Evaluation Metrics} \label{eval_metrics}

\noindent\textbf{Dataset:}
We train SBP exclusively on samples generated by frozen LDMs, without modifying its weights or requiring additional training data. We perform experiments using LDMs trained on CelebA-HQ \cite{karras2017progressive} and FFHQ \cite{karras2019style}. CelebA-HQ contains 30,000 facial images of size \(256\times256\) and is known to exhibit demographic imbalance because it was curated from celebrity images collected from the internet \cite{karkkainenfairface}. FFHQ contains 70,000 high-quality facial images with greater diversity in age, ethnicity, and appearance. For each experiment, the corresponding training dataset is used as the reference set for FID computation. \\

\noindent\textbf{Diffusion Model:}
We use the official pre-trained LDMs \cite{Rombach_2022_CVPR} trained on CelebA-HQ\footnote{\url{https://ommer-lab.com/files/latent-diffusion/celeba.zip}} and FFHQ\footnote{\url{https://ommer-lab.com/files/latent-diffusion/ffhq.zip}}. Both models were trained using a diffusion process with 1000 timesteps. During inference, we employ DDIM sampling with 50 reverse denoising steps, following the standard sampling configuration. We treat the LDM as a white-box model and perform fairness-aware image generation solely through inference-time latent-space manipulation, without modifying its weights or training procedure. \\

\noindent\textbf{Attribute Classifier:}
We predict demographic labels using the FairFace \cite{karkkainenfairface} classifier, a ResNet model trained on the balanced FairFace dataset. Its balanced training distribution helps minimize classifier-induced bias when evaluating generated samples. 

For experiments using the CelebA-HQ LDM, we adopt the binary gender labels provided by CelebA-HQ. Age is divided into three groups: young (10--29 years), adult (30--59 years), and old (60+ years). Race is grouped into four classes: White, Black, Indian, and Asian. We also perform an analysis with a binary race classification, where ``White" remains unchanged and ``Black" includes individuals from both Black and Indian ethnicities, following prior work \cite{parihar2024balancing}. The same demographic protocol is used for experiments with the FFHQ-trained LDM. \\

\noindent\textbf{Evaluation Metrics:}
Perceptual quality is measured using the Fréchet Inception Distance (FID) \cite{heusel2017gans}. Following standard practice, we compute FID\footnote{Official GitHub repository: \url{https://github.com/mseitzer/pytorch-fid}} against an attribute-balanced subset of the corresponding training dataset (CelebA-HQ or FFHQ) used by the pre-trained LDM. Fairness is evaluated using Fairness Discrepancy (FD) \cite{choi2020fair}, defined as the \( \ell_2 \) distance \( ||\mathbf{u} -\mathbf{y}||_2 \) between the predicted demographic distribution \( \mathbf{y} \) and the target uniform distribution \( \mathbf{u} \), where lower values indicate fairer demographic distributions. Computational efficiency is measured using sampling throughput, defined as \( \mathcal{T}=N/\Delta t \), where \(N\) denotes the number of generated samples and \(\Delta t\) is the total generation time. Throughput is hardware-dependent and measured in samples per second (samples/s).

\subsection{Implementation details} \label{sec:impl}

\noindent\textbf{SBP Training:}
To learn semantic decision boundaries, we first generated a synthetic dataset by sampling facial images with the pre-trained LDM and then predicted their demographic attributes. Through empirical analysis, we found that a training set of 15,000 generated samples provides sufficient semantic diversity for reliable boundary learning (see Section \ref{sec:sample_size_train} for details). We extract the corresponding latent representations in the final denoising timestep (\(t=0\)), where semantic information is most discriminative. We used 50\% of the data for training and 50\% for validation during boundary learning.

We implement SBP\footnote{We intend to publicly release the source code and trained SBPs upon acceptance of this paper} as a lightweight single-layer linear classifier operating on the 5,000-dimensional PCA feature subspace. For binary attributes (gender and binary race), we use a single sigmoid output neuron, whereas for multi-class race and age prediction, we use four-way and three-way softmax outputs, respectively. We train all classifiers using stochastic gradient descent (SGD) with a learning rate of \(1\times10^{-3}\), momentum of 0.9, and a weight decay of \(5\times10^{-4}\). All models are trained until convergence, with early stopping applied when the validation accuracy did not improve for 50 consecutive epochs. \\

\noindent\textbf{Inference Pipeline:}
During inference, we apply the learned semantic boundary corresponding to the target demographic attribute to the initial Gaussian latent prior before reverse diffusion begins. Specifically, the unit normal vector of the selected semantic boundary is scaled by a positive guidance strength \(\delta\) and added to or subtracted from the randomly sampled latent code, producing a modified latent representation that serves as input to the frozen diffusion model.

Unless otherwise stated, we use the same inference procedure for every demographic attribute in all experiments. The guidance strength \(\delta \in \mathbb{Z}_{\ge 0}\) controls the magnitude of semantic intervention, where the choice of positive or negative direction (\(\pm\)) determines the side of the learned semantic boundary toward which the sampling process is steered. The two directions therefore increase or decrease the likelihood of generating the corresponding demographic attribute. We select the value of \(\delta\) based on the ablation study presented in Section \ref{sec:exp_ablation}.

We conduct all experiments using PyTorch on a workstation equipped with an NVIDIA A100 GPU with 40\,GB memory. Training the lightweight SBPs incurs negligible computational overhead compared with diffusion model inference.

\subsection{Timestep-Dependent Semantic Structure in LDMs}

\begin{table} [t]
\renewcommand{\arraystretch}{1.2}
\caption{SBP Classifier Accuracy vs. Denoising Timestep}
\label{tab:boundary}
    \centering
    \setlength{\tabcolsep}{2pt}
    \begin{tabular}{ l  c  c  c  c  c} 
    \toprule
    Attributes & \textit{T} = 50 & \textit{T} = 40 & \textit{T} = 25 & \textit{T} = 10 & \textit{T} = 0  \\
    \midrule
    gender & 0.59 & 0.63 & 0.81 & 0.95 & \textbf{0.97} \\
    race (2-class) & 0.71 & 0.77 & 0.88 & 0.91 & \textbf{0.92} \\
    race (4-class) & 0.71 & 0.75 & 0.77 & 0.90 & \textbf{0.91} \\
    age & 0.73 & 0.82 & 0.91 & 0.94 & \textbf{0.95} \\
    \bottomrule
    \end{tabular}
\end{table}

We examine our hypothesis that demographic semantics become progressively more discriminative during reverse denoising by evaluating the semantic separability of latent representations across denoising timesteps. We train an SBP classifier to predict demographic attributes at different denoising timesteps. The classification accuracies are reported in Tab. \ref{tab:boundary}. We observe a clear pattern: prediction accuracy consistently improves as denoising progresses. For example, the gender classification rises from 59\% at (t=50) to 97\% at (t=0), while the four-class race classifier improves from 71\% to 91\%. Similar trends appear for the remaining demographic attributes. This steady increase in classification accuracy indicates that latent representations become increasingly linearly separable during the reverse denoising process.

These findings are consistent with the distribution analysis presented in Fig. \ref{fig:hist-dist}, where the overlap between demographic groups progressively decreases and the corresponding decision boundary becomes more discriminative at later denoising stages. As the latent distributions separate more clearly, the semantic boundary can be estimated with higher confidence, which improves classification accuracy. This observation motivates our decision to learn semantic boundaries from late-stage latent representations, where semantic discriminability is highest.

\begin{figure}[t]
    \centering
    \includegraphics[width=0.75\linewidth]{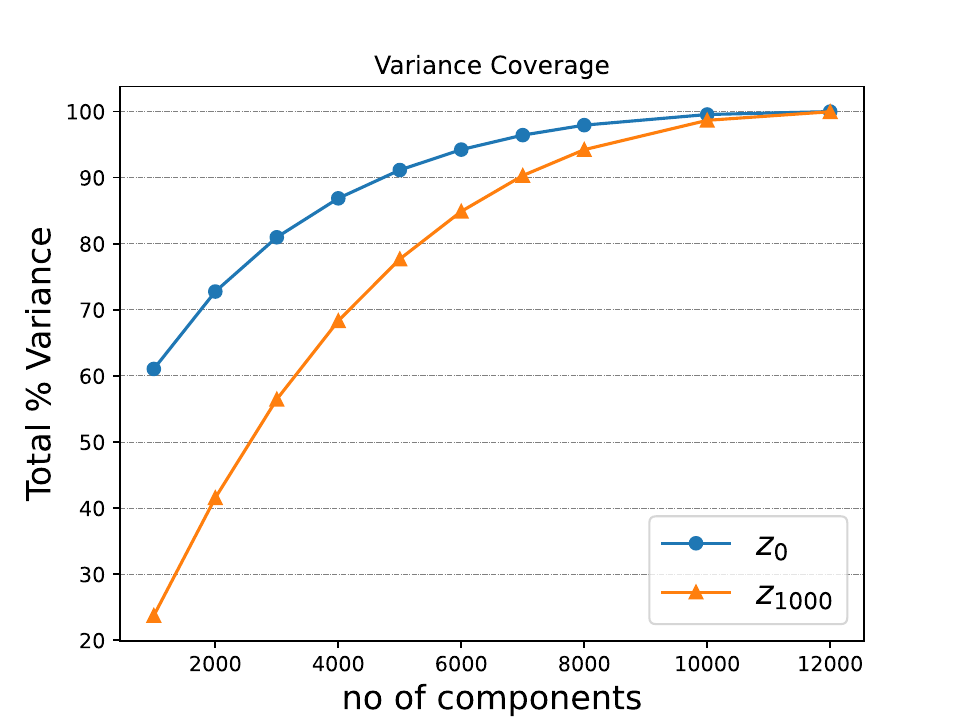}
    \caption{Late-stage latent representations require substantially fewer principal components to preserve the dominant variance than early-stage representations, indicating stronger semantic compactness. This observation supports learning semantic boundaries from late diffusion timesteps in the proposed timestep-decoupled guidance framework.}
    \label{fig:pca}
\end{figure}

\subsection{Analysis of Latent Feature Compactness} \label{sec:exp_pca}
To determine an appropriate feature subspace for semantic boundary learning, we analyze the cumulative variance explained by the principal components extracted from latent representations at different diffusion timesteps. Figure \ref{fig:pca} shows that latent representations in \(t=0\) become highly compact after projection, with approximately 92\% of the variance preserved by 40\% of principal components (5,000 out of 12,288). Increasing dimensionality beyond that point provides only marginal gains while substantially increasing computational cost. We therefore use this subspace in all subsequent experiments.

By contrast, latent representations extracted from the early diffusion stage (\(t=50\)) require substantially more principal components, approximately 8,000, to achieve comparable variance preservation. This indicates that early latent representations remain dominated by noisy and dispersed variation, whereas semantic information becomes progressively concentrated as the reverse diffusion process converges. These observations further support our timestep-decoupled design, in which semantic boundaries are learned from late-stage latent representations that exhibit stronger compactness and semantic separability.

\subsection{Evaluation and Comparisons}
 
We evaluate SBP on five controlled generation tasks: gender, binary race (Black/White), four-class race (White, Black, Asian, Indian), age, and multi-attribute generation (Gender + Race). These attributes were selected since they are the demographic characteristics most commonly studied in fairness-aware face generation and because they span both binary and multi-class semantic-control problems. We compare SBP with three prominent diffusion-based debiasing approaches, Gaussian Harmony \cite{pal2023gaussian}, Unbiased-Diff \cite{perera2025unbiased}, and Balancing Act \cite{parihar2024balancing}, along with the original pre-trained diffusion model under random sampling. Whenever available, we use the quantitative results reported in the original papers. For experiments related to the sampling throughput of Diff-AE and the Balancing Act, and the FID evaluation of the Balancing Act for the gender attribute, we reproduced the corresponding results using the official implementations and the released checkpoints. To ensure a consistent evaluation, we generated 10,000 images per demographic group and assessed their performance using the metrics described in Sec. \ref{eval_metrics}. FID is computed using attribute-wise balanced subsets sampled from the original training distribution. The remainder of this section first evaluates demographic fairness across individual attributes, including gender, race, and age, followed by multi-attribute generation, perceptual quality, and computational efficiency. \\

\begin{table}
\renewcommand{\arraystretch}{1.2}
\caption{Quantitative comparison of demographic fairness (FD\( \downarrow \)) and perceptual quality (FID\( \downarrow \)) for gender generation. Methods are compared with the corresponding pre-trained DM trained on the same dataset.}
\label{tab:compare_gender}
    \centering
    \begin{tabular}{l c c } 
    \toprule
    
    \textbf{Method} & \textbf{FD} \( \downarrow \) & \textbf{FID} \( \downarrow \) \\
    \midrule

    Pre-trained DM (FFHQ) & 0.212 & -  \\
    Gaussian Harmony \cite{pal2023gaussian} & 0.127 & - \\
    \midrule
    
    Pre-trained DM (FFHQ) & 0.109 & 78.44  \\
    Unbiased-Diff \cite{perera2025unbiased} & 0.021 & 78.23  \\
    \midrule

    Pre-trained DM (FFHQ) & 0.068 & 17.45  \\
    SBP (ours) & 0.006 & 18.41 \\
    \midrule

    Pre-trained DM (CelebA) & 0.178 & 54.59  \\
    Balancing Act \cite{parihar2024balancing} & 0.049 & 50.27  \\
    \midrule

    Pre-trained DM (CelebA) & 0.051 & 34.66  \\
    SBP (ours) & 0.001 & 31.54  \\
    \bottomrule
    \end{tabular}
\end{table}

\begin{table} [t]
\renewcommand{\arraystretch}{1.2}
\caption{Quantitative comparison of demographic fairness (FD\( \downarrow \)) and perceptual quality (FID\( \downarrow \)) for racial generation.}
\label{tab:compare_race}
    \centering
    \resizebox{\columnwidth}{!}{%
    \begin{tabular}{l c c c c } 
    \toprule
    \multirow{2}{*}{\textbf{Method}} & \multicolumn{2}{ c }{\textbf{Race (B+W)}} & \multicolumn{2}{ c }{\textbf{Race (4-class)}} \\
    & {FD} \(\downarrow\) & {FID} \(\downarrow\) & {FD} \(\downarrow\) & {FID} \(\downarrow\) \\
    \midrule

    Pre-trained DM (FairFace) & - & - & 0.393 & - \\
    Gaussian Harmony \cite{pal2023gaussian} & - & - & 0.177 & - \\
    \midrule
    
    Pre-trained DM (FairFace) & - & - & 0.222 & 58.98  \\
    Unbiased-Diff \cite{perera2025unbiased} & - & - & 0.014 & 59.24  \\
    \midrule

    Pre-trained DM (CelebA) & 0.334 & 60.01 & 0.292 & 89.14  \\
    Balancing Act \cite{parihar2024balancing} & 0.113 & 52.38 & 0.264 & 91.54  \\
    \midrule

    Pre-trained DM (CelebA) & 0.387 & 39.46 & 0.289 & 34.48  \\
    SBP (ours) & 0.018 & 37.11 & 0.247 & 34.80  \\
    \bottomrule    
    \end{tabular}
    }
\end{table}

\begin{figure*} [t]
    \centering
    \includegraphics[width=0.95\linewidth]{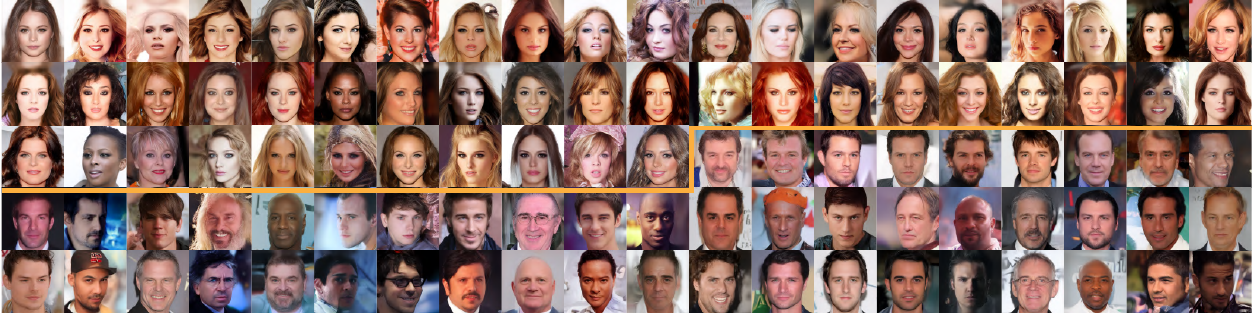}
    \caption{Samples generated with our approach effectively address gender bias by producing an equal distribution of faces: approximately 51\% female above the yellow line and 49\% male below. Additionally, the generated faces exhibit higher perceptual quality, as reflected by the FID score. }
    \label{fig:gender-ours}
\end{figure*}

\noindent\textbf{Gender:}
Gender is one of the most widely studied demographic attributes in fairness-aware face generation because large-scale facial datasets often exhibit noticeable gender imbalance. Previous studies have shown that diffusion models trained on datasets such as CelebA-HQ and FFHQ disproportionately generate female faces during unconditional sampling, making gender an important benchmark for evaluating demographic fairness.

Table \ref{tab:compare_gender} compares SBP with existing diffusion-based debiasing approaches using their respective pre-trained diffusion models. On CelebA-HQ, SBP reduces the fairness disparity from 0.051 to 0.001, corresponding to an approximately 98\% reduction, while also improving the FID from 34.66 to 31.54. The same trend is observed on FFHQ, where SBP reduces the fairness disparity by approximately 91\%, with only a marginal increase in FID.

The reported methods are evaluated on different pre-trained diffusion models, so comparisons should be made within the corresponding training setting. Compared with Gaussian Harmony \cite{pal2023gaussian} and Unbiased-Diff \cite{perera2025unbiased}, both trained on FFHQ, SBP achieves the lowest fairness discrepancy. A direct FID comparison with Gaussian Harmony is not possible since image quality results are not reported. Compared with Balancing Act \cite{parihar2024balancing}, trained on CelebA-HQ, SBP achieves a lower fairness disparity and better image quality.

Figure \ref{fig:gender-ours} presents some examples generated using SBP. We randomly sampled 100 noise samples and applied our decoupled guidance method. Whereas the original LDM predominantly generates female faces during unconditional sampling, SBP produces an approximately balanced distribution (51\% female and 49\% male) while preserving facial diversity and visual realism. These results demonstrate that a single inference-time semantic guidance step is enough to regulate gender representation effectively without modifying the pre-trained diffusion model. \\

\noindent\textbf{Race:}
Race is a more challenging fairness setting than gender because it requires balancing multiple demographic groups at once. To evaluate whether timestep-decoupled guidance generalizes beyond binary semantic control, we consider a four-class race setting with White, Black, Asian, and Indian identities. For direct comparison with earlier work, we also construct a binary race setting (White and Black) following Parihar \etal \cite{parihar2024balancing}, where Brown faces are grouped under the Black category.

We report the quantitative results in Tab. \ref{tab:compare_race}. In the binary race setting, SBP reduces the fairness disparity by approximately 95\% relative to the original CelebA-HQ pre-trained LDM while maintaining a comparable level of image quality. The four-class setting is inherently more difficult because demographic overlap increases as additional groups are considered. Even under this setting, SBP continues to improve demographic balance, achieving a 15\% reduction in fairness disparity while preserving perceptual quality.

The reported methods are evaluated using different pre-trained diffusion models and should therefore be compared within the corresponding experimental setting. Gaussian Harmony \cite{pal2023gaussian} and Unbiased-Diff \cite{perera2025unbiased} use DMs trained on FairFace and report results only for the four-class race setting. Balancing Act \cite{parihar2024balancing} and our method, by contrast, are based on CelebA-HQ and evaluate the binary race task. Within the CelebA-HQ setting, SBP achieves a substantially lower fairness disparity than Balancing Act while maintaining comparable image quality. For the four-class task, although direct comparison across datasets is not appropriate, SBP consistently improves demographic balance over the original pre-trained LDM.

These results demonstrate that timestep-decoupled guidance is not limited to simple binary attributes and generalizes well to more complex demographic distributions. The consistent improvement across the two race settings indicates that the learned semantic boundaries remain informative even as the demographic distribution becomes more complex. \\

\begin{table} [t]
\renewcommand{\arraystretch}{1.2}
\caption{Quantitative comparison of demographic fairness (FD\( \downarrow \)) and perceptual quality (FID\( \downarrow \)) for age generation.}
\label{tab:compare_age}
    \centering
    \begin{tabular}{l l l } 
    \toprule
    \textbf{Method} & \textbf{FD} \(\downarrow\) & \textbf{FID} \(\downarrow\) \\
    \midrule

    Pre-trained DM (CelebA) & 0.256 & 60.68  \\
    Balancing Act \cite{parihar2024balancing} & 0.283 & 71.84  \\
    \midrule

    Pre-trained DM (CelebA) & 0.450 & 46.01  \\
    SBP (ours) & 0.135 & 36.98  \\
    \bottomrule
    \end{tabular}
\end{table}

\noindent\textbf{Age:}
Age differs from gender and race since it forms a gradual semantic continuum rather than a discrete demographic attribute. We report the quantitative comparison in Tab. \ref{tab:compare_age}. Both SBP and Balancing Act \cite{parihar2024balancing} use diffusion models trained on CelebA-HQ, enabling a direct comparison under the same experimental setting. To the best of our knowledge, age has received relatively limited attention in diffusion-based fairness-aware face generation compared with gender and race.

SBP achieves the lowest fairness disparity while preserving image quality. The reduction in fairness disparity is smaller than that observed for gender and binary race, reflecting the increased semantic overlap among the three age groups. Nevertheless, SBP consistently improves demographic balance over both the original pre-trained LDM and Balancing Act, indicating that timestep-decoupled guidance remains effective for attributes with gradual semantic variation. \\

\noindent\textbf{Discussion (RQ1):}
The results in Tables \ref{tab:compare_gender}, \ref{tab:compare_race}, \ref{tab:compare_age} demonstrate that SBP consistently improves demographic fairness across gender, binary race, four-class race, and age while preserving perceptual quality. These findings confirm that a simple semantic boundary predictor learned from latent representations from late-stage is sufficient for effective inference-time demographic control. \\

\begin{table}[t]
\renewcommand{\arraystretch}{1.2}
\caption{Evaluation of FD and FID scores for multi-attribute generation. The best score is highlighted in \textbf{bold}. }
\label{tab:multi}
    \begin{center}
    \begin{tabular}{ l  c  c } 
    \toprule
    Attribute & \multicolumn{2}{ c }{\textbf{Gender + Race}} \\
    Method & {FD} \(\downarrow\) & {FID} \(\downarrow\) \\
    \midrule
    Random Sampling & 0.256 & 60.68 \\
    Balancing Act~\cite{parihar2024balancing} & 0.075 & 49.91  \\
    SBP (ours) & \textbf{0.055} & \textbf{45.16}  \\
    \bottomrule
    \end{tabular}
    \end{center}
\end{table}

\begin{table*} [t]
\renewcommand{\arraystretch}{1.2}
\caption{Evaluation of perceptual image quality (FID scores) for balanced generated samples across various demographics. The best score per metric is emphasized in \textbf{bold}. }
\label{tab:fid}
    \centering
    \makebox[\textwidth]{\begin{tabular}{l | l l | l l l l | l l l } 
    \toprule
    Attribute & \multicolumn{2}{ c |}{\textbf{Gender}} & \multicolumn{4}{ c |}{\textbf{Race}} & \multicolumn{3}{ c }{\textbf{Age}} \\
    Category & {Male} & {Female} & {White} & {Black} & {Indian} & {Asian} & {Young} & {Adult} & {Old} \\
    \midrule
    Random Sampling & 35.14 & \textbf{31.11} & \textbf{30.84} & 58.82 & 76.74 & 72.62 & 71.04 & \textbf{30.13} & 71.11 \\
    Balancing Act~\cite{parihar2024balancing} & 90.02\textsuperscript{\textdagger} & 40.20\textsuperscript{\textdagger} & - & - & - & - & - & - & - \\
    SBP (ours) & \textbf{33.91} & 38.13 & 34.24 & \textbf{45.78} & \textbf{50.60} & \textbf{51.17} & \textbf{48.09} & 32.44 & \textbf{47.03} \\
    \bottomrule
    \end{tabular}}
\end{table*}

\noindent\textbf{Multi-attribute generation:}
Real-world synthetic datasets often require multiple demographic attributes to be balanced simultaneously rather than independently. We therefore evaluate SBP jointly on gender and race generation to test whether the proposed guidance remains effective under compositional semantic control.

We report the quantitative results in Tab. \ref{tab:multi}. Within the CelebA-HQ setting, SBP achieves the lowest fairness disparity (FD = 0.055), corresponding to a 78\% reduction relative to random sampling (FD = 0.256) and a further 26\% reduction compared with Balancing Act (FD = 0.075). At the same time, SBP improves image quality, reducing the FID from 49.91 to 45.16 compared with Balancing Act.

Compared with balancing individual attributes, the fairness disparity naturally increases when gender and race are optimized jointly. For SBP, the FD increases from 0.001 for gender and 0.009 for binary race to 0.055 for joint gender--race generation, reflecting the increased difficulty of simultaneously satisfying multiple demographic constraints. Despite this increase, SBP consistently maintains a more balanced demographic distribution than both the original sampling process and the existing inference-time baseline. \\

\noindent\textbf{Discussion (RQ2):}
The results in Tab. \ref{tab:multi} demonstrate that the proposed semantic guidance strategy generalizes beyond individual demographic attributes, enabling simultaneous control of multiple attributes without retraining or modifying the underlying diffusion model. This finding indicates that the learned semantic boundaries remain effective even under compositional demographic guidance. \\

\noindent\textbf{Fairness versus Image Quality:}
To examine whether demographic balancing affects perceptual quality across demographic groups, we evaluate FID independently of fairness using balanced subsets of CelebA-HQ for each demographic attribute. The corresponding results are reported in Tab. \ref{tab:fid}.

Overall, SBP preserves image quality while improving demographic fairness. For gender, image quality remains comparable to the original LDM, with only modest changes in FID across both categories. Compared with the publicly available Balancing Act checkpoint\textsuperscript{\textdagger}, SBP produces substantially lower FID for both male (33.91 vs. 90.02) and female (38.13 vs. 40.20) faces.

The largest quality improvements are observed for race and age. Averaged across the four racial groups, the FID decreases from 59.76 to 45.45 (approximately 24\%), while the average FID across the three age groups decreases from 57.43 to 42.52 (approximately 26\%). The largest gains occur for previously underrepresented groups, including Indian, Asian, young, and old faces, whereas the dominant White and adult categories exhibit only small changes. This trend suggests that the proposed semantic guidance primarily improves image quality for demographics that are underrepresented in the original diffusion model while preserving the quality of already well-represented groups. \\

\begingroup
\renewcommand{\thefootnote}{\fnsymbol{footnote}}
\footnotetext[2]{The reported values were obtained by evaluating the official checkpoints released by the respective authors.}
\endgroup

\noindent\textbf{Discussion (RQ3):}
The perceptual quality evaluation in Tab. \ref{tab:fid} shows that the proposed inference-time guidance preserves the visual realism of generated images while substantially improving demographic fairness. Larger guidance strengths may modestly increase FID, but an appropriate choice of the guidance strength yields an effective fairness-quality trade-off. \\

\begin{table}
\renewcommand{\arraystretch}{1.2}
\caption{Computational Efficiency Comparison. Sampling throughput was measured in samples/sec. on a single NVIDIA A100.}
\label{tab:comp_time}
    \begin{center}
    \begin{tabular}{ l  l } 
    \toprule
    \multirow{2}{*}{\textbf{Method}}  & {Sampling throughput} \\
                                      & {samples/sec.} \( \uparrow \) \\
    \midrule
    Random Sampling \cite{Rombach_2022_CVPR} & 1.56 \\
    Diff-AE \cite{preechakul2022diffusion} & 0.30 \textsuperscript{\textdagger} \\
    Unbiased-Diff \cite{perera2025unbiased} & 0.26 \\
    Balancing Act \cite{parihar2024balancing} & 1.45 \textsuperscript{\textdagger} \\
    SBP (ours) & 1.41 \\
    \bottomrule
    \end{tabular}
    \end{center}
\end{table}

\noindent\textbf{Computational Efficiency:}
Although fairness is the primary objective, inference-time efficiency is also relevant for practical deployment. The sampling throughput of SBP is compared with representative diffusion-based debiasing methods in Tab. \ref{tab:comp_time}. Throughput is reported at a resolution of \(256 \times 256\) pixels and measured in samples per second on a single NVIDIA A100 GPU. For Diff-AE~\cite{preechakul2022diffusion} and Balancing Act~\cite{parihar2024balancing}, the reported values were obtained using the official checkpoints released by the respective authors\textsuperscript{\textdagger}.

The evaluated methods are based on different diffusion architectures, and the reported throughput should therefore be interpreted within that context. In particular, Unbiased-Diff~\cite{perera2025unbiased} employs a DDPM backbone rather than an LDM, making direct architectural comparisons less meaningful. Its throughput is included for reference.

SBP generates 1.41 samples/sec., compared with 1.56 samples/sec. for the original LDM and 1.45 samples/sec. for Balancing Act. The modest reduction relative to the original model arises from the additional latent-space guidance step applied before reverse denoising. In contrast, Diff-AE reports a substantially lower throughput (0.30 samples/sec.). These results indicate that the proposed guidance introduces little additional inference overhead while maintaining sampling speeds comparable to existing inference-time approaches. \\

\noindent\textbf{Discussion (RQ4):}
The throughput results show that the fairness improvements obtained with SBP require only a modest additional inference cost. SBP performs a single latent-space intervention before reverse diffusion, resulting in sampling speeds that remain close to the original LDM and comparable to existing inference-time guidance methods.

\subsection{Ablation Study}\label{sec:exp_ablation}

During inference, SBP requires only one hyperparameter, the guidance strength \( \delta \), which determines the magnitude of the latent-space shift orthogonal to the learned semantic boundary. Its effect is intuitive: setting a larger value produces a stronger semantic intervention, whereas smaller values yield more conservative adjustments. Since \( \delta \) directly influences both demographic balance and perceptual quality, we first examine its effect. The only additional design choice is the number of latent-attribute pairs used to learn the semantic boundary during SBP training. Together, these two parameters constitute the complete set of tunable components in SBP, making the method straightforward to configure in practice. \\

\begin{table} [t]
\renewcommand{\arraystretch}{1.2}
\caption{Effect of \( \delta \) on Gender Debiasing.}
    \label{tab:ablation_gender}
    \centering
    \setlength{\tabcolsep}{4pt}
    \begin{tabular}{c | c c c c c c}
    \toprule
    Scale \( \delta \) & 0 & 1 & 2 & 3 & 4 & 5 \\
    \midrule
    {FD} \( \downarrow \) & 0.051 & 0.039 & 0.024 & 0.013 & 0.001 & 0.001 \\
    {FID} \( \downarrow \) & 34.66 & 34.78 & 32.52 & 30.17 & 31.54 & 33.2 \\
    \bottomrule
    \end{tabular}
\end{table}

\begin{figure}[b]
    \centering
    \includegraphics[width=1\linewidth]{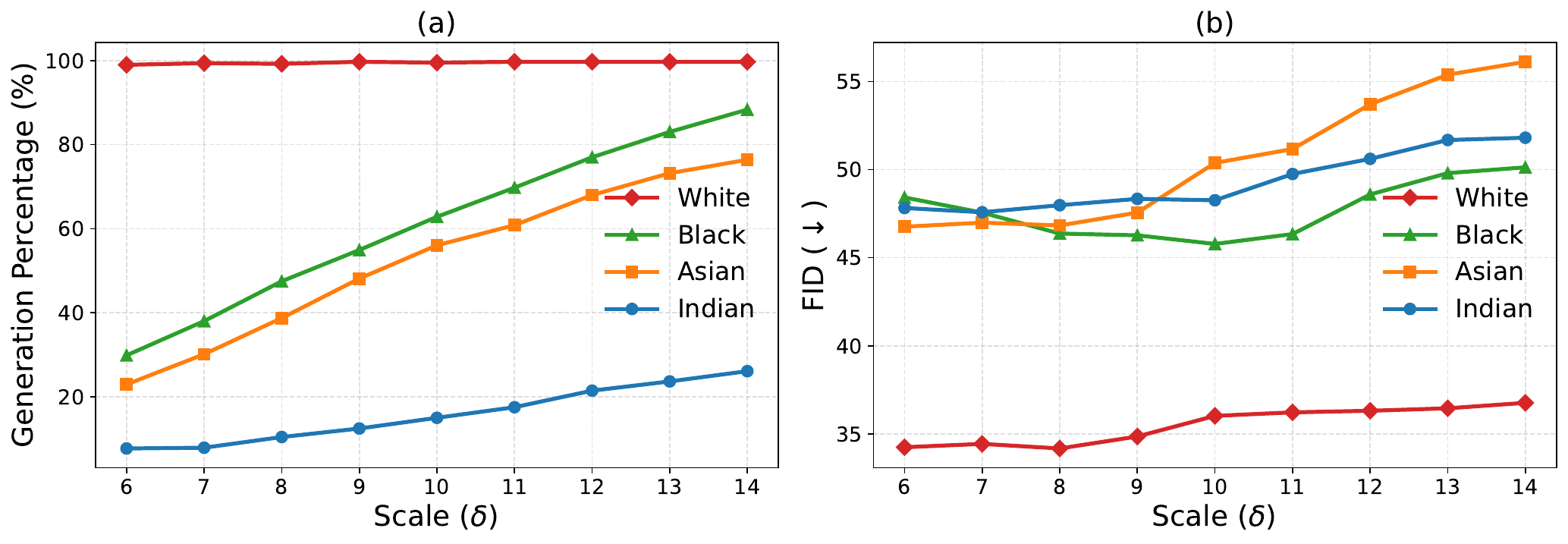}
    \caption{(a) Generation percentage vs. \( \delta \), (b) FID vs. \(\delta \). Effect of the guidance strength \( \delta \) on demographic fairness (FD) and perceptual quality (FID) for four-class race generation. Increasing \( \delta \) strengthens semantic guidance, improving demographic balance while gradually reducing image fidelity beyond the optimal operating point.}
    \label{fig:ablation}
\end{figure}

\noindent\textbf{Effect of Guidance Strength:} \label{apx_sec:scale}
The guidance strength \( \delta \) determines the magnitude of the latent-space perturbation applied orthogonal to the learned semantic boundary before reverse diffusion. Setting \( \delta = 0 \) disables semantic guidance entirely, reducing SBP to the original pre-trained LDM with standard random sampling. Larger values progressively increase the influence of the learned semantic boundary on the generated samples.

We report the effect of varying \( \delta \) for gender generation in Table \ref{tab:ablation_gender}. As \( \delta \) increases from 0 to 4, the fairness disparity decreases from 0.051 to 0.001, corresponding to an approximately 98\% reduction. Over the same range, the FID initially improves from 34.66 to 31.54, indicating that moderate semantic guidance can improve both demographic balance and image quality. Increasing the guidance further (\(\delta=5\)) produces no additional reduction in fairness disparity while increasing the FID from 31.54 to 33.20, suggesting that excessively strong guidance offers little benefit once demographic balance has been achieved.

A similar pattern is observed in four-class race generation, as illustrated in Fig. \ref{fig:ablation}. Increasing \( \delta \) progressively increases the representation of the underrepresented demographic groups, particularly the Black and Asian categories. Beyond intermediate guidance strengths, however, the improvement in demographic balance becomes progressively smaller, while perceptual quality begins to deteriorate. These observations suggest that moderate values of \( \delta \) provide the best balance between fairness and image quality across different demographic attributes. \\

\noindent\textbf{Effect of SBP Training Sample Size:}\label{sec:sample_size_train}
The only training-related design choice in SBP is the number of latent-attribute pairs used to learn the semantic boundary. Table \ref{tab:sample_size} evaluates three training set sizes for the gender attribute.

\begin{table} [t]
\renewcommand{\arraystretch}{1.2}
\caption{Influence of SBP training sample size on the final generation performance measured using FD and FID.}
    \label{tab:sample_size}
    \centering
    \begin{tabular}{ l  c  c }
    \toprule
    Training Set Size & {FD} \(\downarrow \) & {FID} \(\downarrow \) \\
    \midrule
    
    5,000  & 0.002 & 36.27  \\
    15,000 & \textbf{0.001} & 31.54 \\
    45,000 & \textbf{0.001} & \textbf{31.46} \\
    \bottomrule
    \end{tabular}
\end{table}

Increasing the training set from 5,000 to 15,000 samples reduces the fairness disparity from 0.002 to 0.001 while improving the FID from 36.27 to 31.54. Expanding the training set further to 45,000 samples produces only a marginal improvement in FID (31.54 to 31.46), while the fairness disparity remains unchanged. These results indicate that the semantic boundary converges with a relatively modest number of training samples, making SBP inexpensive to train despite operating in a high-dimensional latent space.

\section{Limitations}

The proposed SBP learns a separate semantic boundary for each demographic attribute. Although each boundary can be estimated efficiently from a relatively small number of generated samples, supporting additional attributes requires training and maintaining multiple SBPs. Furthermore, boundary learning depends on external attribute classifiers, so errors or demographic bias in the predicted labels can directly affect the learned semantic boundaries.

Our design is motivated by the distinct semantic roles of the two extreme stages of the denoising process. Accordingly, all experiments learn the semantic boundary from the final denoising timestep and apply it at the initial noisy latent. This strategy consistently improves fairness across all evaluated settings. However, the intermediate timesteps were not examined in this work. 

Future work will investigate whether learning semantic boundaries at intermediate denoising timesteps could yield better transferability or achieve a more favorable balance between demographic fairness and image quality. Another promising direction is to extend the timestep-decoupled semantic guidance beyond demographic attributes to more subtle semantic characteristics.

\section{Conclusion}
In this work, we presented SBP, a one-shot inference-time framework for generating fairness aware samples using latent diffusion models. Our analysis showed that late-stage latent representations are well suited for learning semantic boundaries, whereas early denoising stages provide greater flexibility for semantic guidance. Building on this observation, we introduced a timestep-decoupled semantic guidance strategy that separates semantic boundary learning from semantic intervention across the denoising process. SBP learns semantic boundaries from late-stage latents and transfers them to the initial noisy latent, where a single guidance step steers the reverse diffusion process toward demographically balanced image generation without modifying the remaining denoising trajectory. Experiments on gender, race, and age demonstrate consistent improvements in demographic fairness while preserving perceptual image quality. By combining timestep-decoupled boundary learning with one-shot guidance in a lightweight framework, SBP offers a practical approach for generating balanced synthetic facial data, supporting the development of fairer downstream learning systems without retraining existing latent diffusion models.

\bibliographystyle{unsrt}
\bibliography{references}
\clearpage

\appendix
\section{Latent Diffusion Model} \label{sec:ldm}

Denoising diffusion models (DMs) \cite{ho2020denoising, dhariwal2021diffusion, nichol2021improved} generate images through a forward diffusion process that progressively corrupts data with Gaussian noise and a reverse diffusion process that iteratively removes that noise to recover the data distribution. Latent Diffusion Models (LDMs) \cite{Rombach_2022_CVPR} perform this process in a learned latent space, substantially reducing computational cost while preserving semantic information. \\

\noindent\textbf{Forward Diffusion:}
Given a data sample \( \mathbf{x}_0 \sim q(\mathbf{x}) \), the forward diffusion process is defined as a Markov chain that gradually adds Gaussian noise over \( T \) timesteps according to

\begin{equation} \label{eq:forw}
    q(\mathbf{x}_t | \mathbf{x}_{t-1}) =
    \mathcal{N}\!\left(
    \mathbf{x}_t;
    \sqrt{1-\beta_t}\mathbf{x}_{t-1},
    \beta_t\mathbf{I}
    \right)
\end{equation}

where \( \beta_t \) denotes the noise schedule. \\

\noindent\textbf{Reverse Diffusion:}
Starting from Gaussian noise \( \mathbf{x}_T \sim \mathcal{N}(0,\mathbf{I}) \), the reverse process learns to approximate the posterior \( q(\mathbf{x}_{t-1}|\mathbf{x}_t) \) using a U-Net \cite{ronneberger2015u}. The reverse transition is parameterized as

\begin{equation} \label{eq:rev}
    p_{\theta}(\mathbf{x}_{t-1} | \mathbf{x}_{t}) =
    \mathcal{N}
    \left(
    \mathbf{x}_{t-1};
    \mathbf{\mu}_{\theta}(\mathbf{x}_{t}, t),
    \Sigma_{\theta}(\mathbf{x}_{t}, t)
    \right)
\end{equation}

\noindent\textbf{Training Objective:}
The diffusion model is trained by predicting the injected Gaussian noise using the simplified objective

\begin{equation} \label{eq:dmloss}
    \mathcal{L}_{\textrm{DM}} =
    \mathbb{E}_{\mathbf{x},\epsilon,t}
    \left[
    \left\|
    \epsilon -
    \epsilon_{\theta}(\mathbf{x}_{t},t)
    \right\|_2^2
    \right]
\end{equation}

LDM extends this formulation by performing diffusion in a compressed latent space. An encoder \( \mathcal{E} \) maps an image to its latent representation \( \mathbf{z} \), diffusion operates on \( \mathbf{z} \), and a decoder \( \mathcal{D} \) reconstructs the final image. The corresponding objective is

\begin{equation} \label{eq:ldmloss}
    \mathcal{L}_{\textrm{LDM}} =
    \mathbb{E}_{\mathcal{E}(\mathbf{x}),\epsilon,t}
    \left[
    \left\|
    \epsilon -
    \epsilon_{\theta}(\mathbf{z}_{t},t)
    \right\|_2^2
    \right]
\end{equation}

\section{Intuition for Timestep-Specific Semantic Roles} \label{apx_sec:theory}
We provide an information-theoretic intuition for why semantic boundary estimation and semantic manipulation are naturally suited at different diffusion timesteps. The discussion is intended as a conceptual explanation that complements the empirical findings presented in the main text.

Let \( \mathbf{X} \) denote a generated image, \( \mathbf{Y} \) a demographic attribute of interest (\eg, gender, race, or age), and \( \mathbf{Z}_t \) represent the latent variable at diffusion timestep \( t \). The forward diffusion process induces a Markov chain:
\[
\mathbf{Z}_0 \rightarrow \mathbf{Z}_1 \rightarrow \cdots \rightarrow \mathbf{Z}_T,
\]
where each transition corresponds to a noisy channel. Through inequality in data processing,
\[
I(\mathbf{Z}_t; X) \leq I(\mathbf{Z}_{t-1}; \mathbf{X}), \quad I(\mathbf{Z}_t; \mathbf{Y}) \leq I(\mathbf{Z}_{t-1}; \mathbf{Y}).
\]

In reverse diffusion, denoising does not symmetrically restore all information. Instead, it preferentially reconstructs information necessary for perceptual fidelity, such as texture and fine detail, which may not align with demographic semantics.

Although late-stage latent representations retain substantial semantic information, not all such information is available to be manipulated in a controlled manner. To distinguish semantic information from usable semantic information, we consider the mutual information between a demographic attribute \( \mathbf{Y} \) and a linear projection of the latent representation:
\[
I_{\text{lin}}(\mathbf{Z}_t; \mathbf{Y}) = I(\mathbf{b}_t^\top \mathbf{Z}_t; \mathbf{Y}),
\]
where \( \mathbf{b}_t \) denotes the optimal linear classifier at timestep \( t \). Although it is possible that during the late stages of the diffusion process, the value of \( I(\mathbf{Z}_t; \mathbf{Y}) \) is still high, nonlinear entanglement with perceptual detail causes a decreased value of \( I_{\text{lin}}(\mathbf{Z}_t; \mathbf{Y}) \), restraining the suitability of direct intervention. This is the reason why a high linear separability does not always mean that there can be a high level of manipulability.

From this perspective, we consider timestep selection to be trade-off between two competing quantities: maximizing semantic accessibility, represented by \(I_{\text{lin}}(\mathbf{Z}_t; \mathbf{Y}) \), while minimizing interference with perceptual structure. This trade-off can be expressed as:
\[
t^\star = \arg\max_t \left[ I_{\text{lin}}(\mathbf{Z}_t; \mathbf{Y}) - \lambda I_{\text{per}}(\mathbf{Z}_t; \mathbf{X}) \right],
\]
where \( I_{\text{per}} \) denotes information associated with perceptual detail and \( \lambda \) controls the balance between semantic accessibility and perceptual preservation. Since these objectives are generally not optimized at the same timestep, estimating semantic boundaries and applying semantic guidance need not occur simultaneously.

The proposed timestep-decoupled strategy is based on the hypothesis that coarse semantic structure is preserved along the diffusion trajectory. Rather than assuming that semantic directions remain constant across timesteps, it exploits the observation that diffusion progressively refines image details while maintaining global structure. As a result, semantic boundaries learned from late-stage latent representations can still provide meaningful guidance when transferred to earlier timesteps, where latent modifications propagate more effectively through the subsequent denoising process.

We do not argue the global linearity of demographic attributes or provide a perfect semantic disentanglement. Instead, it provides an intuitive explanation for why semantic information is most discriminative at one stage of the diffusion process yet most amenable to intervention at another, thereby motivating the proposed timestep-decoupled semantic guidance framework.


\section{Timestep-Dependent Semantic Structure in LDMs}
We evaluate linear separability at the two extreme phases by learning the decision boundary in the respective latent space through training the SBP model. \\

\noindent\textbf{Semantic Boundary Estimation at Late Denoising Stages:}

At \( t = 0 \), the latent codes, denoted by \( \mathbf{z}_0 \), encode well-formed semantics of facial images. We build a dataset by pairing these latent codes with attribute labels \( y \) predicted from the corresponding generated images. A linear classifier trained on such data provides highly accurate decision boundaries on demographic features.

\begin{figure}[t]
\centering
\includegraphics[width=1\linewidth]{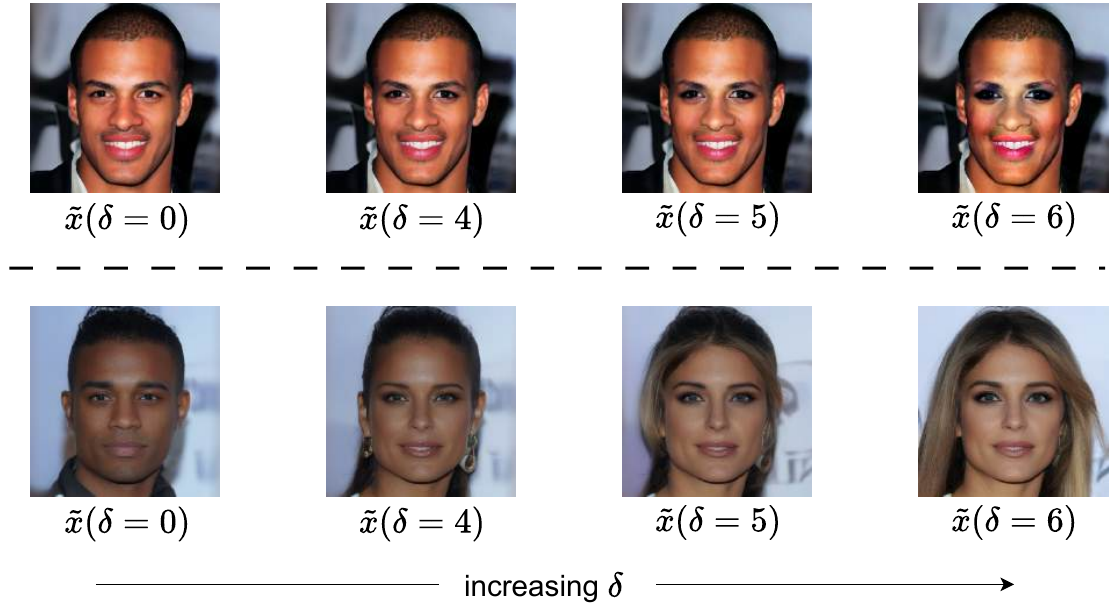}
\caption{The top row depicts image manipulation with the latent representations when \( t = 0 \), that is, at the stage of well formed facial semantics. Attribute changes mainly affect fine-grained features around the lips and eyes, like glossy lips or eyes with kajal. The bottom row shows the manipulation at \( t=T \) in which semantic separability is less, but transformations yield visually consistent results through coarser structural changes.}
\label{fig:difft}
\end{figure}

Despite the strong separability, manipulating \( \mathbf{z}_0 \) along the normal vector for the gender attribute results in limited global changes because this stage focuses on refining complex details. Modifications at this stage mainly affect fine-grained regions such as the lips and eyes, as illustrated in the top row of Fig. \ref{fig:difft}. This behavior reflects the role of late-stage denoising, which focuses on refining complex and interdependent visual details, making it difficult to shift demographic attributes once a well-formed facial structure has already taken shape. \\

\noindent\textbf{Early-Stage Latent Intervention:}

In contrast, latent representations at \( t = T \) lack explicit semantic structure, leading to lower SBP accuracy. However, perturbations at this stage yield perceptually coherent transformations, as early denoising steps predominantly influence coarse facial geometry \cite{choi2022perception}. As shown in the bottom row of Fig. \ref{fig:difft}, traversing semantic directions at this timestep produces more noticeable attribute changes. However, since the semantic separation in the latent space is not well-defined at this stage, the generated images may fail to maintain a balanced attribute distribution for a given demographic. This suggests that while early-stage manipulation facilitates significant transformations, it does not preserve the structured separability required for controlled attribute generations.

In order to effectively integrate these complementary properties, we adopt a timestep-decoupled strategy which decomposes the two mechanisms of acquiring semantic boundaries, on the one hand, and the use of semantic intervention, on the other.

Our strategy assumes that coarse semantic alignment is preserved across diffusion timesteps, so that the directional biases that were introduced during the initializing step can affect the denoising in successive timesteps. This assumption is supported by the observation that denoising trajectories maintain global semantic structure while progressively refining perceptual detail.

\section{Ablation Study} \label{apx_sec:ablation}

\subsection{The Impact of Sample Size Generation on Bias } \label{apx_sec:sample_size_bias}

Since demographic fairness is estimated from generated samples, it is important to verify that the measured bias is not an artifact of the evaluation sample size. We therefore evaluated gender bias using 2,000, 5,000, and 10,000 generated images. As shown in Tab. \ref{tab:sample_fd_fid}, the measured fairness disparity remains largely unchanged across different sample sizes, indicating that the evaluation is stable. Based on this observation, we use 10,000 generated samples in all our experiments, consistent with Balancing Act \cite{parihar2024balancing}.

\begin{table} [t]
\renewcommand{\arraystretch}{1.2}
\caption{Impact of Sample Size on Gender Bias in Generated Samples.}
    \label{tab:sample_fd_fid}
    \centering
    \begin{tabular}{ c  c  c }
    \toprule
    no. of samples & {FD} \( \downarrow \) & {FID} \( \downarrow \) \\
    \midrule
    
    2,000  & 0.001 & 34.13  \\
    5,000  & 0.001 & 32.49 \\
    10,000 & 0.001 & 31.54 \\
    \bottomrule
    \end{tabular}
\end{table}

\subsection{Qualitative Analysis of Semantic Guidance} \label{sec:delta_gender}

\begin{figure*}
    \centering
    \includegraphics[width=0.9\linewidth]{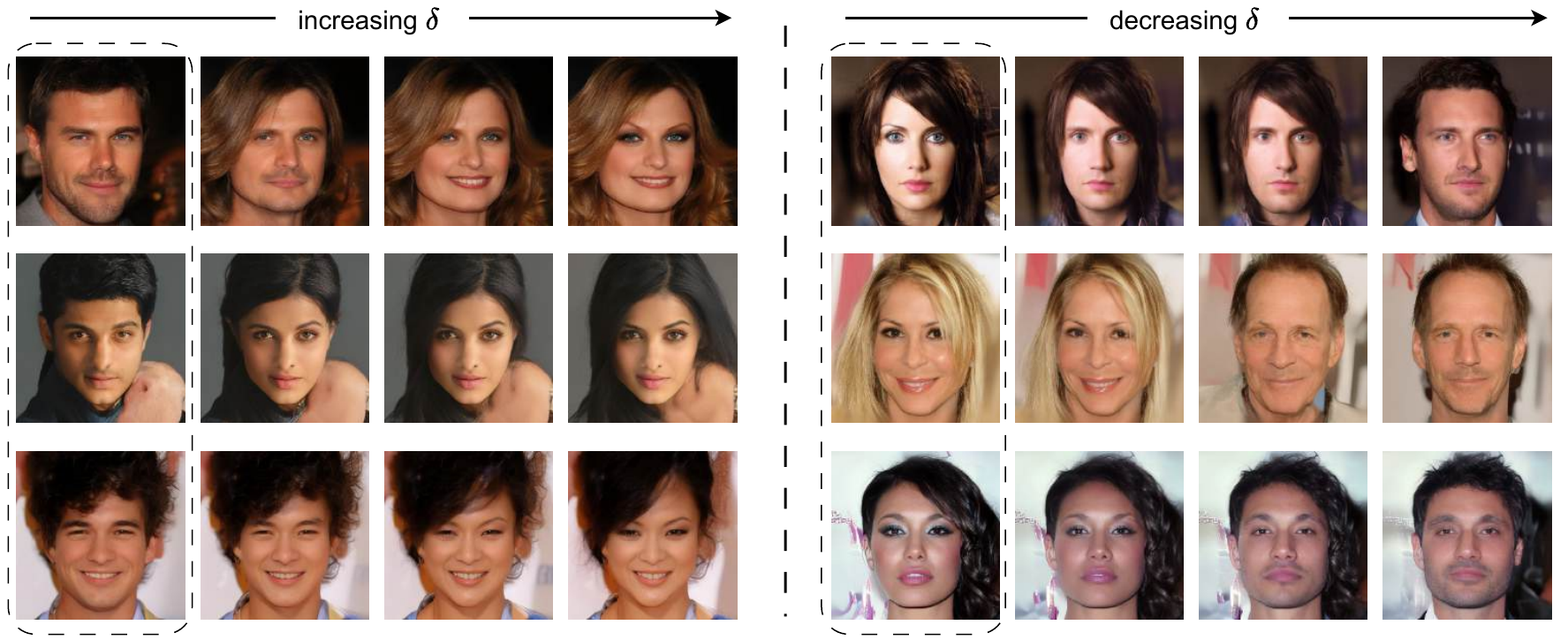}
    \caption{Qualitative visualization of gender guidance. The guidance strength \( \delta \) controls movement along the learned semantic boundary. Increasing \( \delta \) progressively changes male faces to female faces (left), whereas decreasing \( \delta \) produces the reverse transition (right), while preserving image quality.}
    \label{fig:delta_gender}
\end{figure*}

Figure \ref{fig:delta_gender} illustrates the effect of traversing the learned gender semantic boundary. Adding the latent vector scaled by \( (\delta) \) progressively transforms male faces into female faces, whereas subtracting the same latent vector produces the opposite transition. The smooth progression demonstrates that the learned boundary captures a meaningful semantic direction in the latent space and enables controllable demographic manipulation through a single inference-time guidance step.

\section{Qualitative Results} \label{apx_sec:results}

We present additional qualitative results demonstrating the effectiveness of the proposed SBP. We compare unconditional samples generated by the original Latent Diffusion Model with those obtained using SBP across different demographic attributes, highlighting changes in demographic representation while assessing the preservation of perceptual image quality. Figures are best viewed in color at 2\(\times\)--4\(\times\) zoom.

\begin{figure*}[ht] 
        \centering
        \includegraphics[width=1\linewidth]{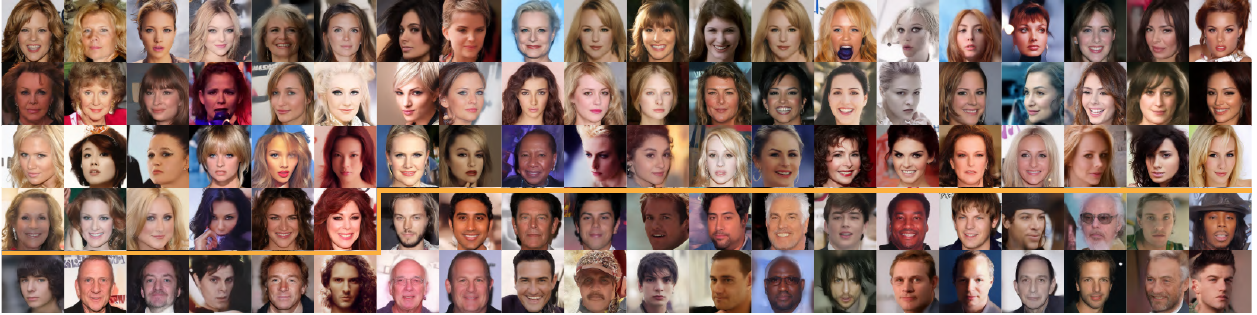}
        \caption{Samples from the LDM \cite{Rombach_2022_CVPR} model trained on CelebA-HQ, demonstrating the model's tendency to produce disproportionately female faces (66\%) shown above the yellow line and male faces (34\%) below, amplifying the gender bias found in the original data.}
        \label{fig:gender-ldm}
\end{figure*}

\begin{figure*} [t]
    \centering
    \includegraphics[width=1\linewidth]{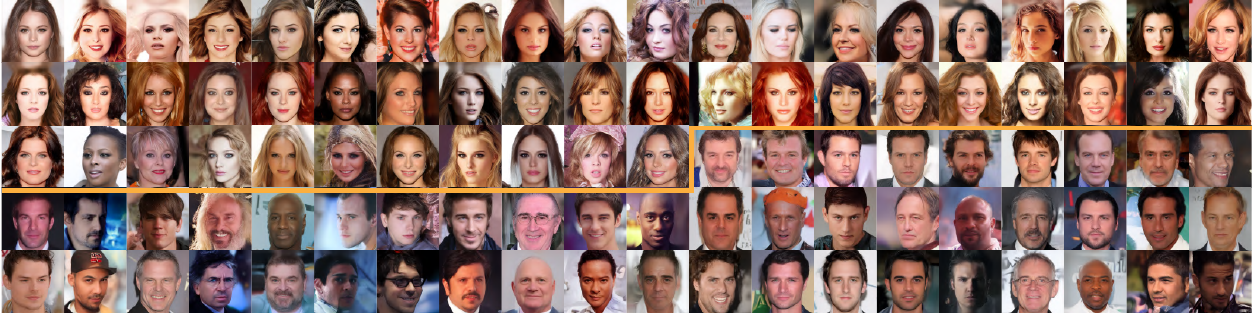}
    \caption{Samples generated with our approach effectively address gender bias by producing an equal distribution of faces: approximately 51\% female above the yellow line and 49\% male below. Additionally, the generated faces exhibit higher perceptual quality, as reflected by the FID score. }
    \label{fig:gender-ours2}
\end{figure*}

Fig. \ref{fig:gender-ldm} illustrates the gender bias inherent in unconditional sampling from a Latent Diffusion Model trained on CelebA-HQ, where generated faces are dominated by female subjects. In contrast, applying the proposed Semantic Boundary Predictor substantially rebalances the demographic distribution by increasing the representation of under-represented groups, while preserving the perceptual quality of the generated faces as observed in Fig. \ref{fig:gender-ours2}.

\begin{figure*}[ht]
    \centering
    \includegraphics[width=1\linewidth]{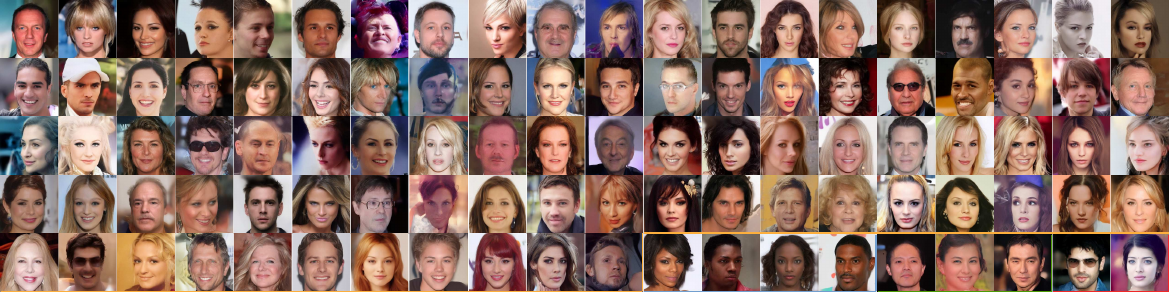}
    \caption{Samples from the LDM \cite{Rombach_2022_CVPR} model trained on CelebA-HQ, highlighting the model's inherent bias towards generating White faces (91\%), bounded by yellow line, disproportionately. Black faces (4\%) bounded by the blue line, Asian faces (3\%) bounded by the green line, and Indian faces (2\%) bounded by the red line are underrepresented, reflecting the racial bias present in the original dataset.}
    \label{fig:race_ldm}
\end{figure*}

\begin{figure*}[ht]
    \centering
    \includegraphics[width=1\linewidth]{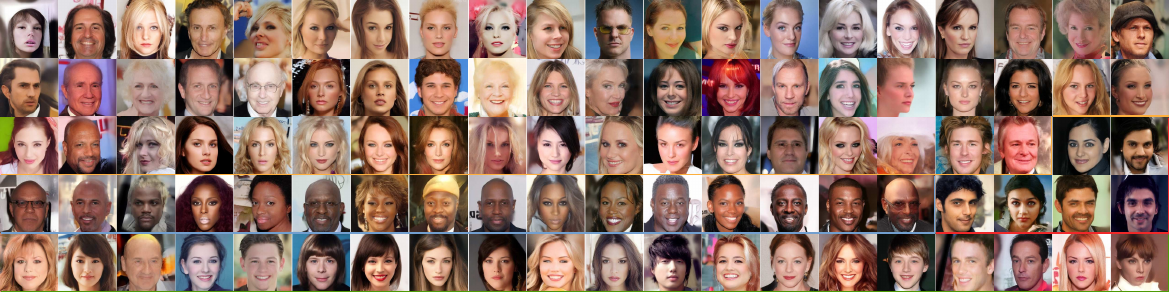}
    \caption{Our LDM + SBP approach mitigates racial bias by upsampling underrepresented groups. The generated samples demonstrate a more balanced distribution: White faces (58\%) bounded by the yellow line, Black faces (16\%) bounded by the blue line, Asian faces (20\%) bounded by the green line, and Indian faces (6\%) bounded by the red line. Despite the upsampling, the generated faces maintain high perceptual quality.} 
    \label{fig:race_ours}
\end{figure*}

Fig. \ref{fig:race_ldm} illustrates the racial bias inherent in unconditional sampling from a Latent Diffusion Model trained on CelebA-HQ, where generated faces are dominated by white subjects and under-represent other racial groups. In contrast, applying the proposed Semantic Boundary Predictor substantially rebalances the demographic distribution by increasing the representation of under-represented groups, while preserving the perceptual quality of the generated faces as observed in Fig. \ref{fig:race_ours}. 

\end{document}